\def\year{2019}\relax

\documentclass[letterpaper]{article} 
\usepackage{aaai19}  
\usepackage{times}  
\usepackage{helvet}  
\usepackage{courier}  
\usepackage{url}  
\usepackage{graphicx}  
\nocopyright
\usepackage{times}
\usepackage{helvet}
\usepackage{courier}
\usepackage{CJKutf8}
\usepackage{amsmath}
\usepackage{bm}
\usepackage{algorithm}
\usepackage[mathscr]{eucal}
\usepackage{bbm}
\usepackage{multirow,multicol}
\usepackage{xcolor}
\usepackage{algpseudocode} 
\usepackage{flushend}
\usepackage{caption}
\usepackage{subcaption}
\usepackage{graphicx}  
\usepackage{xpinyin}
\begin{document}
\begin{CJK}{UTF8}{gbsn}

\title{Improving the Robustness of Speech Translation}

\author{
Xiang Li$^{1,2}$\thanks{Work done while at Sogou Inc.},
Haiyang Xue$^{1,2,*}$\thanks{Corresponding Author},
Wei Chen$^{3}$,
Yang Liu$^{4}$,
Yang Feng$^{1,2}$,
Qun Liu$^{5}$\\  
$^1$Institute of Computing Technology, Chinese Academy of Sciences \\
$^2$University of Chinese Academy of Sciences\\
$^3$Voice Interaction Technology Center, Sogou Inc., Beijing\\
$^4$Department of Computer Science and Technology, Tsinghua University, Beijing\\
$^5$Huawei Noah's Ark Lab, Huawei Technologies, Hong Kong\\
lixiang@ict.ac.cn, xuehaiyang@ict.ac.cn, chenweibj8871@sogou-inc.com,\\
liuyang2011@tsinghua.edu.cn, fengyang@ict.ac.cn, qun.liu@huawei.com
}

\maketitle
\begin{abstract}
Although neural machine translation (NMT) has achieved impressive progress recently, it is usually trained on the clean parallel data set and hence cannot work well when the input sentence is the production of the automatic speech recognition (ASR) system due to the enormous errors in the source. To solve this problem, we propose a simple but effective method to improve the robustness of NMT in the case of speech translation. We simulate the noise existing in the realistic output of the ASR system and inject them into the clean parallel data so that NMT can work under similar word distributions during training and testing. Besides, we also incorporate the Chinese Pinyin feature which is easy to get in speech translation to further improve the translation performance. Experiment results show that our method has a more stable performance and outperforms the baseline by an average of 3.12~\textsc{BLEU} on multiple noisy test sets, even while achieves a generalization improvement on the WMT'17 Chinese-English test set.
\end{abstract}

\section{Introduction}

In recent years, neural machine translation (NMT) has achieved impressive progress and has outperformed statistical machine translation (SMT) systems on multiple language pairs~\cite{sennrich2016wmt}. NMT models are usually built under the encoder-decoder architecture where the encoder produces a representation for the source sentence and the decoder generates target translation from this representation word by word~\cite{cho2014learning,sutskever2014sequence,vaswani2017attention}.

Despite its success, NMT is sensible to the orthographic errors that human beings can comprehend as expected~\cite{belinkov2017synthetic}. This problem is aggravated in speech translation where the output of the automatic speech recognition (ASR) system is used as the input of NMT which usually contains more noise. The example in Table~\ref{table:asr_noise_cases} shows that the conventional NMT system fails to translates the misrecognized ASR output correctly. It is reported that the increase of error rate of ASR brings a significant performance degradation of machine translation~\cite{le2017disentangling}. It indicates that the best NMT systems have to observe a performance decline due to high ASR error rate under noisy environmental conditions, such as simultaneous interpretation, even though ASR has matured to the point of commercial applications. Conventional NMT systems are usually trained on the high-quality written parallel data which hardly contain many ASR-specific errors, resulting in a mismatch between training data and test data. An ideal solution is to train NMT systems on training data in the form of erroneous speech transcriptions paired with their counterpart translations. Unfortunately, this kind of corpora available is somewhat scarce and expensive to collect. Therefore, in addition to reducing the error rate of ASR, it is necessary to improve the robustness of NMT to the inevitable ASR errors.

\begin{table}[t]
\centering
\begin{tabular}{l|l}
\textbf{Speech} & \xpinyin*{这}  \xpinyin*{份} \xpinyin*{礼} \xpinyin*{物}   \xpinyin*{饱} \xpinyin*{含} \xpinyin*{一}  \xpinyin*{份}  \textbf{\color{blue}\xpinyin*{深}} \textbf{\color{blue}\xpinyin*{情}} \\ \hline 
\textbf{ASR} & \xpinyin*{这}  \xpinyin*{份} \xpinyin*{礼} \xpinyin*{物}   \xpinyin*{饱} \xpinyin*{含}  \xpinyin*{一}  \xpinyin*{份}  \textbf{\color{red}\xpinyin*{申}} \textbf{\color{red}\xpinyin*{请}} \\ \hline
\textbf{Reference} & This gift is full of affection \\ \hline
\textbf{NMT} & This gift contains an application \\
\end{tabular}
\caption{\label{table:asr_noise_cases} An example of speech translation. For this example, the original word ``深情''(highlighted by blue color) which means affection is misrecognized as its homophonic word ``申请''(highlighted by red color) which means application, leading to an inaccurate translation generated by the conventional NMT system.}
\end{table}

In this paper, our goal is to improve the robustness of NMT to erroneous ASR outputs in the speech translation scenario. We propose an effective and computationally inexpensive approach to craft a large number of ASR-specific noise training examples by simulating realistic ASR errors, in order to alleviate the problem of insufficient speech translation training data. The basic idea is to randomly substitute some correct source tokens by other noise tokens at each training iteration. And we propose four strategies of choosing ASR-specific noise symbols. Using our approach, it is easy to obtain a robust NMT model only based on the standard NMT training method without modifying the training objective and extra computation load of introducing generative adversarial networks with a training difficulty~\cite{arjovsky2017wasserstein,cheng2018towards}.

To achieve a further improvement in translation quality, it is desirable to recover the inherent semantic relationship among source characters which is broken by introduced ad-hoc noise at training time. For this purpose, in addition to the standard character-level representation, we also propose to explicitly incorporate the syllable-level representation (also called Pinyin\footnote{The official romanization system for Mandarin Chinese.}) of a Chinese character as an additional input feature, resulting in a novel Pinyin-aware embedding of Chinese characters.  

We conduct experiments on WMT'17 Chinese-English translation task. Experimental results show that our approaches not only significantly enhance the robustness of NMT on the artificial noisy test sets, but also improve the generalization performance of NMT on the original test set. We finally illustrate the advantage and disadvantage of our robust NMT system via two real-world examples.

\section{The Challenge of Speech Translation}

The dominated speech translation systems generally employ a cascaded architecture which consists of an ASR component followed by an NMT component. An ASR system ingests user utterances as inputs and generates text transcriptions as outputs. And then an NMT system consumes these transcriptions and produces translations in another language.

Recently, there has been growing interest in building an end-to-end ASR system as a way of folding separate acoustic, pronunciation, and language modeling components of a conventional ASR system into a single neural network~\cite{chiu2017state}. We consider Listen, Attend and Spell (LAS)~\cite{chan2016listen} as an example to formally describe the basic principle of end-to-end ASR. The basic LAS model consists of three sub-modules: the listener, the attender and the speller. Let $\mathcal{D}_{\mathbf{x},\mathbf{z}}=\big\{\langle \mathbf{x}^{(n)},\mathbf{z}^{(n)} \rangle\big\}^{N}_{n=1}$ be our training data of LAS. For the $n$-th instance, $\mathbf{x}_{n}={x}_{1},\ldots,x_{t},\ldots,x_{T}$ is the input sequence of filter bank spectra features, and $\mathbf{z}^{(n)}={z}_{1},\ldots,{z}_{l},\ldots,z_{L}$ is the output sequence of transcriptions. The listener maps $\mathbf{x}^{(n)}$ to a high-level feature representation $\mathbf{H}$. The attender takes $\mathbf{H}$ and determines which listener features in $\mathbf{H}$ should be attended to predict the next output symbol $z_{l}$. Finally, the speller accepts the output of the attender in order to produce a probability distribution $P(z_{l}|z_{<l},\mathbf{x}^{(n)})$. The standard training objective is to find a set of model parameters that minimizes the negative log-likelihood on the ASR training data:
\begin{equation}\label{asr}
\begin{aligned}
\hat{\bm{\theta}}_{A}=\mathop{\arg\min}_{\bm{\theta}_{A}}\Big\{-\sum\limits_{n=1}^{N}\log{P\big(\mathbf{z}^{(n)}|\mathbf{x}^{(n)};\bm{\theta}_{A}\big)}\Big\},
\end{aligned}
\end{equation}
where $\bm{\theta}_{A}$ is a set of ASR model parameters, and $z_{<l}=(z_{1},\ldots,z_{l-1})$ is the sequence of previous symbols.

Given a bilingual written training data $\mathcal{D}_{\mathbf{z},\mathbf{y}}=\big\{\langle \mathbf{z}^{(m)},\mathbf{y}^{(m)} \rangle\big\}^{M}_{m=1}$. For the $m$-th sentence pair, let $\mathbf{z}^{(m)}={z}_{1},\ldots,{z}_{i},\ldots,z_{I}$ be our source-language sequence, and $\mathbf{y}^{(m)}={y}_{1},\ldots,y_{j},\ldots,y_{J}$ be our target-language sequence. NMT usually models the translation probability as 
\begin{equation}\label{eq1}
\begin{aligned}
P(\mathbf{y}^{(m)}|\mathbf{z}^{(m)};\bm{\theta}_{N})=\prod^{J}_{j=1}P(y_{j}|\mathbf{z}^{(m)},y_{<j};\bm{\theta}_{N}),
\end{aligned}
\end{equation}
where $\bm{\theta}_{N}$ represents a set of NMT model parameters, and $y_{<j}={y}_{1},\ldots,y_{j-1}$ is a partial translation. The probability of generating the $j$-th target token is usually calculated as
\begin{equation}\label{eq2}
\begin{aligned}
P(y_{j}|\mathbf{z}^{(m)},y_{<j};\bm{\theta}_{N})=\mathrm{softmax}\big(g(\mathbf{y}_{j-1}, \mathbf{h}_{j}, \mathbf{c}_{j}, \bm{\theta}_{N})\big),
\end{aligned}
\end{equation}
where $\mathbf{y}_{j-1}$ is the word embedding of ${y}_{j-1}$, $\mathbf{h}_{j}$ is a hidden state at $j$-th step, $\mathbf{c}_{j}$ is a vector representing the context on source side for generating the $j$-th target word, and $g(\cdot)$ is a non-linear activation function. The standard training objective is to find a set of model parameters that minimizes the negative log-likelihood on the training data:
\begin{equation}\label{eq3}
\begin{aligned}
\hat{\bm{\theta}}_{N}=\mathop{\arg\min}_{\bm{\theta}_{N}}\Big\{-\sum\limits_{m=1}^{M}\log{P\big(\mathbf{y}^{(m)}|\mathbf{z}^{(m)};\bm{\theta}_{N}\big)}\Big\}.
\end{aligned}
\end{equation}

\begin{table}[t]
\centering
\begin{tabular}{l|c|ccccc}
Error type   & Rate  & \multicolumn{5}{c}{Example} \\ \hline
Substitution & 6.4\% & \xpinyin*{语}  & \xpinyin*{音} & \xpinyin*{翻} & \textbf{\color{red}\xpinyin*{一}} & \\ \hline
Deletion     & 2.3\% &         & \xpinyin*{音} & \xpinyin*{翻} & \xpinyin*{译}	 &  \\ \hline
Insertion    & 0.7\% & \xpinyin*{语}  & \xpinyin*{音} & \xpinyin*{翻} & \textbf{\color{red}\xpinyin*{了}}
\end{tabular}
\caption{\label{table:wer}Error rates of three error categories for ASR. For the speech input ``语音翻译'' which means speech translation, ``译'' is substituted by ``一'' in the first case, ``音'' is deleted in the second case, and ``了'' is inserted in the third case.}
\end{table}

Since the source side of training data in NMT does not match the ASR outputs, the performance of the NMT system is adversely affected by the ASR system which is prone to recognition errors due to the regional accents of speakers or environmental noise conditions.

Once the erroneous ASR system is deployed, a wise way to improve the quality of speech translation is to adapt the downstream NMT system to ASR errors which are generally classified into three categories (i.e.\ substitution, deletion, and insertion) based on their Levenshtein alignments between the transcription and its reference. We provide three examples to illustrate these ASR error categories in Table~\ref{table:wer}. We also investigate the word error rate (WER) of our in-house Chinese ASR system on our in-house evaluation dataset which consists of about 100 hours of Chinese speech cross multiple domains. As shown in Table~\ref{table:wer}, the substitution error rate (6.4\%) becomes the majority of WER (9.4\%). Our observation is consistent with prior work~\cite{mirzaei2016automatic}. And It is known that over 50\% of the machine translation errors are associated with substitution errors which have a greater impact on translation quality than deletion or insertion errors~\cite{vilar2006error,ruiz2014assessing}. Hence, our goal is to improve the robustness of NMT to the substitution errors. 

\section{Approach}

In this section, we propose four strategies to craft ASR-specific noise training examples. To further improve the translation quality of Chinese-sourced NMT, we propose to incorporate the Chinese Pinyin as an additional input feature. 

\subsection{Which characters to be substituted}

Determining which correct characters could be substituted becomes a prerequisite for crafting noise examples. Inspired by \textit{dropout}~\cite{hinton2012improving,srivastava2014dropout,gal2016theoretically}, we propose to randomly substitute some source characters of parallel data with minor noise that the conventional NMT system is not able to translate correctly with high confidence, in order to simulate the substitution errors of ASR and regularize the NMT model.

For each source sentence $\mathbf{z}$ of NMT training data, we posit a vector $\bm{r}$ with $|\mathbf{z}|$ independent Bernoulli random variables, each of which has substitution probability $p$ of being 1. The vector is sampled and multiplied element-wise with the character ID inputs of the input layer in NMT, to creates the distorted training examples $\tilde{\mathbf{z}}$ by substituting the remaining characters labeled with 1 according to Equation~\ref{equation:substitute}. The distorted training examples are then used as input to the input layer.

\begin{equation}\label{bernoulli}
\begin{aligned}
\bm{r}_{c}\sim \textnormal{Bernoulli}(p)
\end{aligned}
\end{equation}
\begin{equation}\label{equation:substitute}
\begin{aligned}
\tilde{\mathbf{z}}=
\begin{cases}
\tilde{c} & \text{ if } \bm{r}_{c}=1 \\ 
c & \text{ if } \bm{r}_{c}=0,
\end{cases}
\end{aligned}
\end{equation}
where $\tilde{c}$ is a noise symbol.

In this case, the original $\mathcal{D}_{\mathbf{z},\mathbf{y}}$ is perturbed into $\mathcal{D}_{\tilde{\mathbf{z}},\mathbf{y}}$ of which the source side shares similar distribution with ASR outputs. Therefore, the robustness of NMT can be improved by observing a large number of variants of ASR-specific noise examples without changing the standard training method. 

\subsection{How to choose noise}

We design four noising schemes of sampling noise symbols to substitute the determined source positions:
\begin{itemize}
\item \textbf{Placeholder-based Substitution}

We first propose a simple and general method to only consider the special placeholder ``\textless{}SUB\textgreater'' which hardly appears in the wild as the noise symbol. Our motivation is that forcing the model to reduce character dependencies. Using this approach, our NMT model theoretically should observe $2^{|\textbf{z}|}$ variants for each source sentence $\textbf{z}$.

\item \textbf{Uniform Distribution-based Substitution}

Since the placeholder ``\textless{}SUB\textgreater'' hardly appears in the realistic ASR outputs, there is still a mismatch between the perturbed training data and ASR outputs. We propose to substitute the source positions with a sampled noise from the uniform distribution described in Equation~\ref{equation:uniform}.

\begin{equation}\label{equation:uniform}
\begin{aligned}
P(\tilde{c})=\frac{1}{|\mathcal{V}|},
\end{aligned}
\end{equation}
where $\mathcal{V}$ is the source vocabulary.

\begin{table}[t]
\centering
\begin{tabular}{l|cccc}
\textbf{Methods}   & \multicolumn{4}{c}{\textbf{Noise Example}} \\ \hline
Placeholder     & \xpinyin*{语} & \textbf{\color{red}\textless{}SUB\textgreater} & \xpinyin*{翻} & \xpinyin*{译} \\ \hline
Uniform   		& \xpinyin*{语} & \textbf{\color{red}\xpinyin*{饕}} & \xpinyin*{翻} & \xpinyin*{译} \\ \hline
Frequency 		& \xpinyin*{语} & \textbf{\color{red}\xpinyin*{好}} & \xpinyin*{翻} & \xpinyin*{译} \\ \hline
Homophone 		& \xpinyin*{语} & \textbf{\color{red}\xpinyin*{因}} & \xpinyin*{翻} & \xpinyin*{译}                    
\end{tabular}
\caption{\label{table:sampling}Example of our noise sampling methods. For the original source sentence ``语音翻译'', the second character is randomly picked to be substituted by noise (highlighted by red color). For Placeholder method, ``音'' is substituted with the placeholder ``\textless{}SUB\textgreater''. For Uniform method, since each character has equal probability to be noise, ``音'' is substituted with ``饕'' which is a low-frequency character in Chinese. But for Frequency method, ``音'' is substituted by ``好'' which is a high-frequency Chinese character. Finally for Homophone, ``音'' is substituted by ``因'', both the characters share the same pronunciation.}
\end{table}

\item \textbf{Frequency-based Substitution}
 
It is well known that it is difficult for ASR systems to recognize infrequent tokens in the training data~\cite{goldwater2010words}. In other words, the tokens with low frequency in the utterance tend to be misrecognized as frequent tokens. To simulate the real-world ASR scenario, we propose to substitute the source positions with a sample from the following unigram frequency distribution:
\begin{equation}\label{freq-based}
\begin{aligned}
P(\tilde{c})=\frac{Count(\tilde{c})}{\sum\limits_{\tilde{c}'\in \mathcal{V}\backslash \left\{c\right\}}Count(\tilde{c}')},
\end{aligned}
\end{equation}
where $Count$ is a function used to calculate the character frequency in the training data.

\item \textbf{Homophone-based Substitution}

For ASR outputs, another important fact is that there are significant possibilities that a character is substituted by its homophones which pronounce the same as the original one but differ in meaning~\cite{li2008information}. Therefore, we propose to substitute the source positions with a sample of their homophone vocabulary according to the following distribution:
\begin{equation}\label{homo-based}
\begin{aligned}
P(\tilde{c})=\frac{Count(\tilde{c})}{\sum\limits_{\tilde{c}'\in \mathcal{V}(c)\backslash \left\{c\right\}}Count(\tilde{c}')},
\end{aligned}
\end{equation}
where $\mathcal{V}(c)$ is a vocabulary where each character shares the same pronunciation with $c$.

\end{itemize}

Using the crafted noise training examples, the model is forced to learn the more general representation of that perturbed training data and to allocate output stability on the classes of simulated errors.

\subsection{Chinese Pinyin-aware Input Embeddings}

Using the proposed methods of crafting noise examples, the source part of training data is randomly corrupted with minor confusing characters constantly at training time. It indicates that the distorted characters are rare during the whole training process, leading to an issue of data sparsity. 

Because Chinese is famous for its numerous homophones, more than half of the Chinese Internet homophones retain the same pronunciation as their base words~\cite{tang2014study}. When a person types a word on a keyboard, he encounters more than one variants of characters of the word, so users choose a malapropism, which is an incorrect word in place of a word with a similar sound, to express their intense emotions. For example, a web-user picks the word ``\xpinyin*{砖家}'' instead of ``\xpinyin*{专家}'' which is the meaning of specialist, since both the words share the same pronunciation, but the former itself is another ironic name of ``specialist'' who specializes in talking nonsense in the Chinese Internet language. In this case, representing each Chinese character only by their surface symbols intuitively implies that any pair of characters is as distinct as any other pair. This ignores any common Pinyin sequences shared by characters. However, human beings generally have no obstacle to understanding this kind of informal or inaccurate Chinese text as long as the pronunciation is correct. Therefore, it gives us a hint that Pinyin information is helpful to generalize knowledge learned about a character to another via their shared Pinyin sequences since the Chinese syllable level constraints are not as restrictive as surface character sequences.

However, many Chinese characters share the same Pinyin without tones yet not their meanings. For example, ``砖'' which means brick and ``专'' which means specific. By considering a surface character and its Pinyin as equivalent, the performance of NMT models can be harmed by this new source of ambiguity. Due to this concern, we propose to apply a factored input embeddings by combining both character and Pinyin representations motivated by~\citeauthor{sennrich2016linguistic}~\shortcite{sennrich2016linguistic}.

Given a Pinyin sequence $\textbf{p}=p_{1},\ldots,p_{i},\ldots,p_{I}$ which has the same length as the character sequence $\textbf{z}=z_{1},\ldots,z_{i},\ldots,z_{I}$, we look up separate embedding vectors for character and pinyin, and the final factored input embedding $\mathrm{e}_{i}$ for each position $i$ can be generated by concatenating the character embedding $\mathrm{E}_\mathrm{c}[z_{i}]$ and Pinyin embedding $\mathrm{E}_\mathrm{p}(p_{i})$ as
\begin{equation}\label{factor_nmt}
\begin{aligned}
\mathrm{e}_{i}=[\mathrm{E}_\mathrm{c}[z_{i}];\mathrm{E}_\mathrm{p}[p_{i}]],
\end{aligned}
\end{equation}
where $\mathrm{E}_\mathrm{c}$ and $\mathrm{E}_\mathrm{p}$ indicate the feature embedding matrices of character and Pinyin respectively. $\mathrm{e}_{i}$ is actually feed to the encoder of NMT model instead of $\mathrm{E}_\mathrm{c}[x_{i}]$, and all other components of the NMT model remain unchanged.

\begin{table}[t]
\begin{tabular}{l|llll}
\textbf{Input Feature}    &     \multicolumn{4}{c}{\textbf{Example}} \\ \hline
Character	& \multicolumn{1}{c}{\xpinyin*{语}} & \multicolumn{1}{c}{\xpinyin*{音}}  & \multicolumn{1}{c}{\xpinyin*{翻}}  & \multicolumn{1}{c}{\xpinyin*{译}} \\ \hline
Pinyin		& \multicolumn{1}{c}{yu} & \multicolumn{1}{c}{yin} & \multicolumn{1}{c}{fan} & \multicolumn{1}{c}{yi} \\ \hline
Factored	& \multicolumn{1}{c}{[语;yu]} & \multicolumn{1}{c}{[音;yin]} & \multicolumn{1}{c}{[翻;fan]} & \multicolumn{1}{c}{[译;yi]}                     
\end{tabular}
\caption{\label{table:pinyin}Examples of character, Pinyin, and factored input features of the sentence ``语音翻译''.}
\end{table}

\section{Experiments}

\subsection{Setup}

We conduct all experiments on the WMT'17 Chinese-English translation task. The training data consists of 9.3M bilingual sentence pairs obtained by combining the CWMT corpora and News Commentary v12. We use newsdev2017 and newstest2017 as our validation set and clean test set, respectively. Due to lack of public datasets for speech translation, we craft three noisy test sets with different amount of homophones errors in order to simulate the homophonic substitution errors of ASR. And we construct three noisy variants for each source sentence of newstest2017 to increase the diversity of noisy characters. Therefore, the size of each artificial noisy test set is three times larger than newstest2017. We argue that the setup is very close to the realistic speech translation scenario. 

It is well known that NMT benefits from the increasing amount of training data~\cite{koehn2017six}. In addition to WMT training data, we also evaluate the best performing system on our in-house large-scale Chinese-English training data with about 80M sentence pairs. 

\begin{table*}[htb]
\centering
\begin{tabular}{l|c|c|cccc}
\multirow{2}{*}{System} & \multirow{2}{*}{$p$} & \multirow{2}{*}{Clean} & \multicolumn{4}{c}{Noise} \\ \cline{4-7}
                        &                                    &                                 & 1 Sub     & 2 Subs    & 3 Subs    & Ave.    \\ \hline
Baseline                                         & -    & 22.62        & 21.02     & 19.60     & 18.67     & 19.76    \\
~~~~+Pinyin                                          & -    & 22.69        & 21.28     & 19.97     & 18.88     & 20.04    \\ \hline \hline
\multicolumn{1}{l|}{
\multirow{3}{*}{Placeholder}} 					 & 0.1  & 23.10        & 21.82     & 20.76     & 19.66     & 20.75    \\
\multicolumn{1}{l|}{}                            & 0.2  & 23.17        & 21.96     & 20.82     & 19.83     & 20.87 	  \\
\multicolumn{1}{l|}{}                            & 0.3  & 22.54        & 21.51     & 20.61     & 19.63     & 20.58    \\ \cline{2-7} 
\multirow{3}{*}{~~~~+Pinyin} 					 & 0.1  & 22.99        & 21.43     & 22.05     & 20.90     & 21.46    \\
\multicolumn{1}{l|}{}                            & 0.2  & 23.37        & 22.83     & 22.55     & 21.96     & 22.45    \\
\multicolumn{1}{l|}{}                            & 0.3  & 23.31        & 22.88     & 22.40     & 22.10     & 22.46    \\ \hline \hline
\multirow{3}{*}{Uniform}                         & 0.1  & 23.04        & 21.97     & 20.69     & 19.98     & 20.88    \\
                                                 & 0.2  & 23.26        & 22.04     & 20.93     & 20.05     & 21.01    \\
                                                 & 0.3  & 22.85        & 21.71     & 20.74     & 19.92     & 20.79    \\ \cline{2-7} 
\multirow{3}{*}{~~~~+Pinyin} 					 & 0.1  & 23.20        & 22.45     & 21.16     & 20.32     & 21.31    \\
\multicolumn{1}{l|}{}                            & 0.2  & 23.04        & 22.47     & 21.37     & 20.77     & 21.54    \\
\multicolumn{1}{l|}{}                            & 0.3  & 23.07        & 22.31     & 21.30     & 20.47     & 21.36    \\ \hline \hline                                                 
\multirow{3}{*}{Frequency}                       & 0.1  & 23.21        & 22.63     & 22.48     & 21.49     & 22.19	  \\
                                                 & 0.2  & 23.02        & 22.41     & 21.90     & 21.35     & 21.89    \\
                                                 & 0.3  & 21.70        & 21.09     & 20.65     & 20.50     & 20.75    \\ \cline{2-7}
\multirow{3}{*}{~~~~+Pinyin} 					 & 0.1  & \textbf{23.41}        & 22.56     & 22.11     & 20.97     & 21.88    \\
\multicolumn{1}{l|}{}                            & 0.2  & 23.35        & 22.74     & 22.22     & 21.27     & 22.08    \\
\multicolumn{1}{l|}{}                            & 0.3  & 23.11        & 22.52     & 21.81     & 20.82     & 21.72    \\ \hline \hline                                               
\multirow{3}{*}{Homophone}                       & 0.1  & 23.04        & 22.64     & 22.31     & 21.56     & 22.17    \\
                                                 & 0.2  & 23.00        & 22.50     & 22.11     & 21.69     & 22.10    \\
                                                 & 0.3  & 23.07        & 22.78     & 22.43     & 22.24     & 22.48    \\ \cline{2-7}
\multirow{3}{*}{~~~~+Pinyin} 					 & 0.1  & 23.20        & \textbf{23.07}     & \textbf{22.98}     & 22.59     & \textbf{22.88    }\\
\multicolumn{1}{l|}{}                            & 0.2  & 23.06        & 22.88     & 22.89     & \textbf{22.78}     & 22.85    \\
\multicolumn{1}{l|}{}                            & 0.3  & 22.31        & 22.22     & 22.41     & 22.26     & 22.30                                                                                                    
\end{tabular}
\caption{\label{table:exp1}Case-sensitive BLEU scores of our approaches on the clean test set (newstest2017) and three artificial noisy test sets (1 Sub, 2 Subs and 3 Subs) which are crafted by randomly substituting one, two and three original characters of each source sentence in the clean test set with their homophones, respectively. $p$ is the substitution rate. ``Placeholder'' means the placeholder ``$\langle$SUB$\rangle$'' is used as the noise token. ``Uniform'' indicates the uniform distribution based noise sampling. ``Frequency'' represents character frequency based noise sampling. ``Homophone'' denotes Chinese homophone based noise sampling. ``Pinyin'' means incorporating the Chinese Pinyin as an additional input feature.}
\end{table*}

All of the following experiments are carried out based on the Transformer~\cite{vaswani2017attention}, which is similar to conventional NMT models except depending entirely on self-attention and position-wise, and uses fully connected layers for both the encoder and decoder instead of recurrent neural networks or convolutions.

We set the size of all input and output layers to 512 and that of inner-FFN layer to 2048. Training is performed on a single server with 8 Nvidia M40 GPUs. We use a batch size of 4096 on each GPU containing a set of sentence pairs with approximately 4096 source tokens and 4096 target tokens. We train each model with the sentences of length up to 100 words in the training data. We train each model for a total of 600K steps and save the checkpoints with an interval of 1000 training steps. We use a single model obtained by averaging 20 checkpoints that perform best on the development set as the final model for testing. During decoding, we set the beam size to 4. Other training parameters are the same as the default configuration of the Transformer base model.

We report case-sensitive NIST BLEU~\cite{papineni2002bleu} scores for all the systems. For evaluation, we first merge output tokens back to their untokenized representation using~\textit{detokenizer.pl} and then use \textit{mteval-v13a.pl} to compute the scores as per WMT reference.

In this work, we focus on crafting ASR-specific noise examples and incorporating the Chinese Pinyin feature to improve the robustness of NMT. Therefore, we consider the $\mathcal{L}_\mathrm{noisy}$ loss function proposed by~\citeauthor{cheng2018towards}~\shortcite{cheng2018towards} as our training objective. We will omit an exhaustive background description of the loss function and refer readers to~\citeauthor{cheng2018towards}~\shortcite{cheng2018towards}. It is worth noting that our approach can be applied with other adversarial training methods proposed by~\citeauthor{cheng2018towards}~\shortcite{cheng2018towards}.

\subsection{Robustness Performance}\label{sec:rate}

Our character substitution has a hyper-parameter $p\in [0,1]$ which means the probability of substituting a character in the inputs. In this section, we explore the effect of tuning this hyper-parameter. 

The results from Table~\ref{table:exp1} shows that both the Placeholder and Uniform models work best at $p=0.2$, $p=0.1$ is optimal for the Frequency model, and the Homophone model achieves the best performance at $p=0.3$. It indicates that different noise sampling methods have their own optimal substitution rate. Hence, it is hard to set a universal substitution rate for all the models. It also can be seen that the Homophone model behaves stably on all noisy test sets even the substitution rate increases. However, in the case of more noise, other models suffer more performance degradation. We suspect that the homophone noise which still keeps the latent semantic information does not hinder the training process severely.

\subsection{Translation Performance}

Although \textit{dropout} is used for full-connected layers in all models, the baseline model still fails to translate the noise inputs. The results in Table~\ref{table:exp1} show that the baseline model degrades significantly on the test set ``1 SUB'', and the performance becomes worse as noise increases in the other noisy test sets. It also demonstrates that the conventional NMT is indeed fragile to permuted inputs, which is consistent with prior work~\cite{belinkov2017synthetic,cheng2018towards}.

However, our methods make the NMT model more robust to noise inputs. First, the simple Placeholder model achieves an improvement of translation quality over the baseline model from +0.94~\textsc{BLEU} to +1.16~\textsc{BLEU} as the amount of homophone noise characters increases from 1 to 3, according to the results in Table~\ref{table:exp1}. Therefore, it proves that randomly substituting some characters of inputs is a simple yet effective regularizer for conventional NMT. We also evaluate the performance of the Uniform model which uses Chinese characters as substitutions. The results in Table~\ref{table:exp1} suggest that the Uniform model achieves an improvement over the Placeholder model marginally. Then it can be seen that the Frequency model not only significantly enhances the robustness of NMT over the baseline system, but also improves further over the Uniform model up to an average of +1.18~\textsc{BLEU} on the noisy test sets. Compared with the Uniform model, the improvement of Frequency model is especially substantial for noise text with more than one incorrect character. Finally, we can find that the Homophone model performs best and achieved a significant improvement on noise text over the baseline model up to +2.72~\textsc{BLEU}.  

We observe that all our robustness-enhanced models outperform the baseline model on the clean test set up to +0.63~\textsc{BLEU}. And the translation performance of the Homophone model on the ``1 Sub'' is also superior to that one of the baseline model on the clean test. Moreover, even on the ``3 Subs'' with three noise characters, the performance degradation of the Homophone model is only -0.83~\textsc{BLEU}, while the baseline model falls up to -3.95~\textsc{BLEU}.

\subsection{Pinyin Feature}

\begin{figure}[tbp]
\centering
\includegraphics[width=0.5\textwidth]{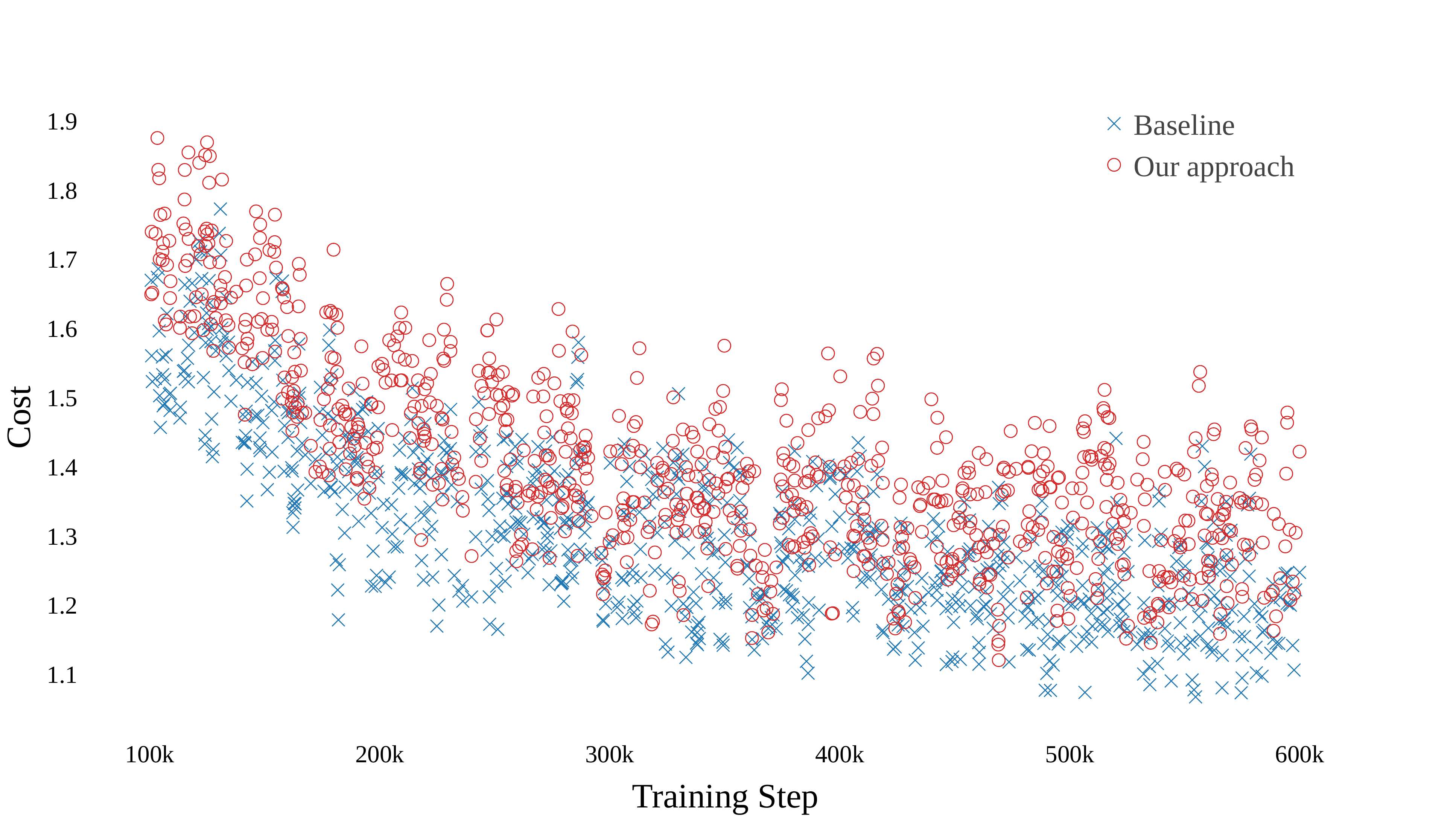}
\caption{Training cost of the baseline model and our robust system.}
\label{fig:cost}
\end{figure}

In this section, we evaluate the performance of our method incorporated with Chinese Pinyin feature. We use the ChineseTone\footnote{https://github.com/letiantian/ChineseTone} tool to convert Chinese characters into their Pinyin counterpart without tones. For the sake of a fair comparison, we keep the total size of input embedding fixed to 512 by setting the embedding sizes of character and Pinyin to 64 and 448, respectively for each system with Pinyin.

As shown in Table~\ref{table:exp1}, Chinese Pinyin feature provides further robustness improvements for the baseline system on all the noisy test sets. It also can be seen that the Homophone model with Pinyin feature achieves a further improvement by an average of +0.71~\textsc{BLEU} on the noisy test sets and a slight generalization improvement on the clean test set. It demonstrates that Pinyin is an effective input feature for improving the robustness of Chinese-sourced NMT.

\begin{table*}[htb]
\centering
\begin{tabular}{l|c|c|cccc}
\multirow{2}{*}{System} & \multirow{2}{*}{Training data size} & \multirow{2}{*}{Clean} & \multicolumn{4}{c}{Noise} \\ \cline{4-7} 
                        &                                    &                                 & 1 Sub     & 2 Subs    & 3 Subs    & Ave.    \\ \hline
\multirow{2}{*}{Baseline}   & ~~9M  & 22.62        & 21.02    & 19.60    & 18.67    & 19.76 \\
                            & 80M & 26.30        & 24.28    & 22.47    & 20.55    & 21.77 \\ \hline \hline
\multirow{2}{*}{Our Approach} & ~~9M  & 23.20        & 23.07    & 22.98    & 22.83    & 22.96 \\
                            & 80M & 26.10        & 25.76    & 25.67    & 25.56    & 25.68
\end{tabular}
\caption{\label{table:exp2}Effect of training data size.}
\end{table*}

\begin{table*}[t]
\centering
\begin{tabular}{l|l}
\textbf{Speech} & \xpinyin*{该} \textbf{\color{blue}\xpinyin*{数}} \xpinyin*{字} \xpinyin*{已} \xpinyin*{经} \xpinyin*{大} \xpinyin*{幅} \xpinyin*{下} \xpinyin*{滑} \xpinyin*{近} 90\% \\ \hline
\textbf{ASR} & \xpinyin*{该} \textbf{\color{red}\xpinyin*{书}} \xpinyin*{字} \xpinyin*{已} \xpinyin*{经} \xpinyin*{大} \xpinyin*{幅} \xpinyin*{下} \xpinyin*{滑} \xpinyin*{近} 90\% \\ \hline
\textbf{Ref}      & The figure has fallen sharply by almost 90\%          \\ \hline
\textbf{Baseline}       & The book has fallen by nearly 90\%          \\ \hline
\textbf{Our approach}	& The figure has fallen by nearly 90\% \\
\end{tabular}
\caption{\label{table:robust_cases} For the same erroneous ASR output, translations of the baseline NMT system and our robust NMT system.}
\end{table*}

It is worth noting that the Placeholder model with Pinyin feature achieves a significant improvement over the original Placeholder model on noisy test sets up to +1.59~\textsc{BLEU}. We suspect that Pinyin feature effectively compensates the model for lost semantic information at training time.

Among all our models, the Homophone model with Pinyin feature achieves a comparable performance on the clean test set, but performs best on the noisy test sets. It suggests that the Homophone model achieves a tradeoff between robustness and generalization. Therefore, the Homophone model with substitution rate 0.1 is used as the best performing NMT model in the subsequent experiments.

\subsection{Training Cost}

We also investigate the training cost of our robust system and the baseline system. As shown in Figure~\ref{fig:cost}, it is obvious that the training cost of baseline model is lower than that one of our robust system during the training process, but our robust system achieves a higher \textsc{BLEU} score. It indicates that our approach effectively improves the generalization performance of the conventional NMT model trained on clean training data.

\subsection{Effect of Source Sentence Length}

\begin{figure}[tbp]
\centering
\includegraphics[width=0.5\textwidth]{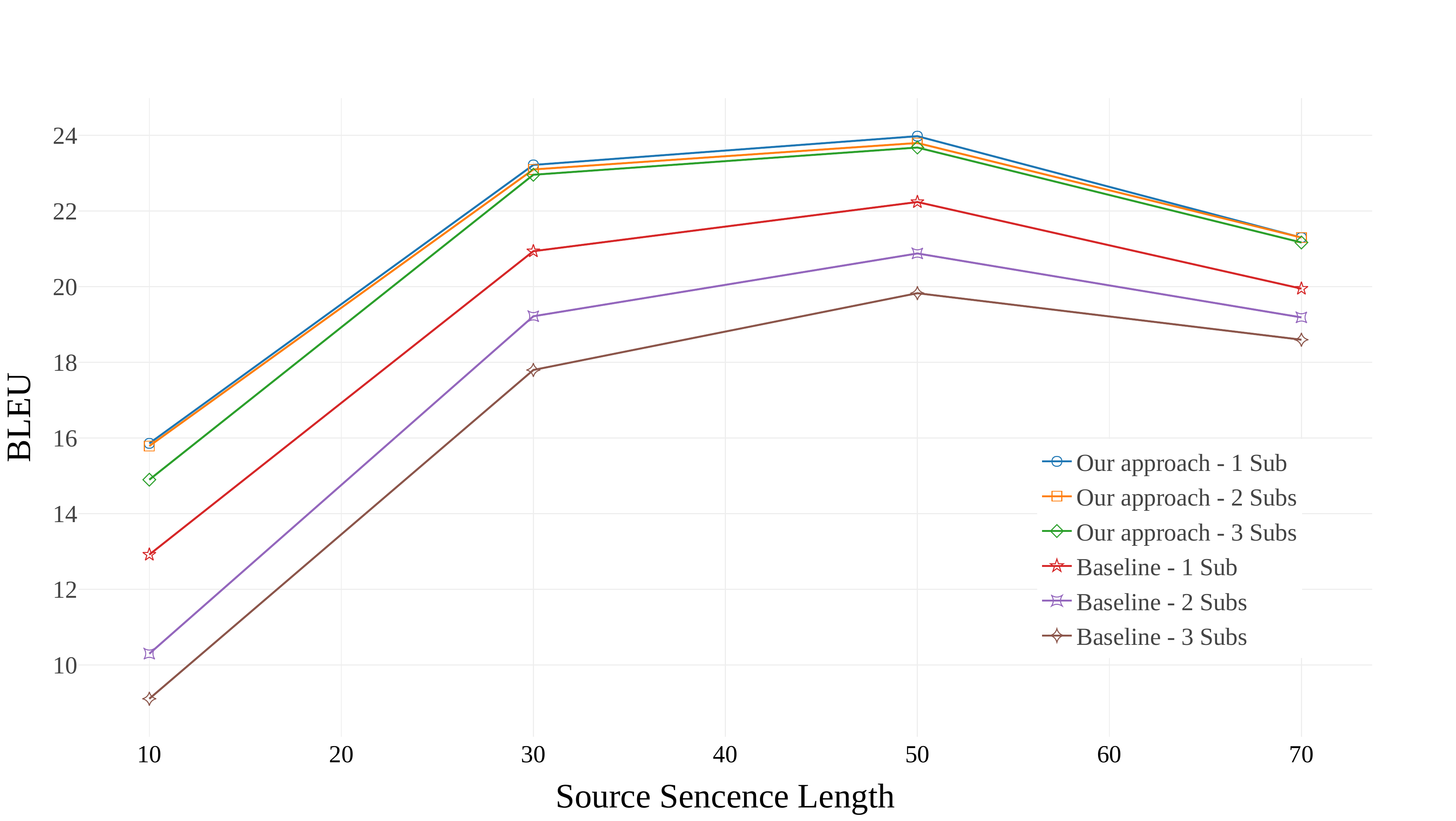}
\caption{Effect of source sentence lengths of noisy input.}
\label{fig:length}
\end{figure}

We also evaluate the performance of our robust system and the baseline on the noisy test sets with different source sentence lengths. As shown in Figure~\ref{fig:length}, the translation quality of both systems is improved as the length increases and then degrades as the length exceeds 50. Our observation is also consistent with prior work~\cite{bahdanau2014neural}. It implies that more context is helpful to noise disambiguation. It also can be seen that our robust system outperforms the baseline model on all the noisy test sets.

\subsection{Effect of Training Data Size}

As shown in Table~\ref{table:exp2}, increasing training data significantly improves the baseline system up to 3.68~\textsc{BLEU} on the clean test data, but only achieves a robustness improvement by an average of +2.01~\textsc{BLEU} on the noisy test sets. It demonstrates that the degradation of translation quality caused by noise is still unavoidable for the conventional NMT model even trained on a larger scale of training data. In contrast, our robust system achieves a comparable improvement on the noisy test sets to the performance on the clean data (2.72~\textsc{BLEU} vs.\ 2.9~\textsc{BLEU}). It shows that our method is stable and effective to NMT regardless of the amount of training data. Compared with the baseline system, it also can be seen that more training data brings more robustness improvement for our robust system on the noisy data (2.72~\textsc{BLEU} vs.\ 2.01~\textsc{BLEU}). It presents that our method can make better use of a larger amount of training data to enhance the robustness of NMT further.

\subsection{A Case Study}

In Table~\ref{table:robust_cases}, we provide a realistic example to illustrate the advantage of our robust NMT system on erroneous ASR output. For this case, the syntactic structure and meaning of the original sentence are destroyed since the original character ``数'' which means digit is misrecognized as the character ``书'' which means book. ``数'' and ``书'' share the same pronunciation without  tones. Human beings generally have no obstacle to understanding this flawed sentence with the aid of its correct pronunciation. The baseline NMT system can hardly avoid the translation of ``书'' which is a high-frequency character with explicit word sense. In contrast, our robust NMT system can translate this sentence correctly. We also observe that our system works well even if the original character ``数'' is substituted with other homophones, such as ``舒'' which means comfortable. It shows that our system has a powerful ability to recover the minor ASR error. We consider that the robustness improvement is mainly attributed to our proposed ASR-specific noise training and Chinese Pinyin feature.

\section{Related Work}

It is necessary to enhance the robustness of machine translation since the ASR system carries misrecognized transcriptions over into the downstream MT system in the SLT scenario. Prior work attempted to induce noise by considering the realistic ASR outputs as the source corpora used for training MT systems~\cite{peitz2012spoken,tsvetkov2014augmenting}. Although the problem of error propagation could be alleviated by the promising end-to-end speech translation models~\cite{serdyuk2018towards,berard2018end}. Unfortunately, there are few training data in the form of speech paired with text translations. In contrast, our approach utilizes the large-scale written parallel corpora. Recently,~\citeauthor{sperber2017neural}~\shortcite{sperber2017neural} adapted the NMT model to noise outputs from ASR, where they introduced artificially corrupted inputs during the training process and only achieved minor improvements on noisy input but harmed the translation quality on clean text. However, our approach not only significantly enhances the robustness of NMT on noisy test sets, but also improves the generalization performance.

In the context of NMT, a similar approach was very recently proposed by~\citeauthor{cheng2018towards}~\shortcite{cheng2018towards}, where they proposed two methods of constructing adversarial samples with minor perturbations to train NMT models more robust by supervising both the encoder and decoder to represent similarly for both the perturbed input sentence and its original counterpart. In contrast, our approach has several advantages: 1) our method of constructing noise examples is efficient yet straightforward without expensive computation of words similarity at training time; 2) our method has only one hyper-parameter without putting too much effort into performance tuning; 3) the training of our approach performs efficiently without pre-training of NMT models and complicated discriminator; 4) our approach achieves a stable performance on noise input with different amount of errors.

Our approach is motivated by the work of NMT incorporated with linguistic input features~\cite{sennrich2016linguistic}. Chinese linguistic features, such as radicals and Pinyin, have been demonstrated effective to Chinese-sourced NMT~\cite{zhang2017improving,du2017pinyin} and Chinese ASR~\cite{chan2016online}. We also incorporate Pinyin as an additional input feature in the robust NMT model, aiming at improving the robustness of NMT further.

\section{Conclusion}

Erroneous ASR is a challenge to speech translation. We propose a simple yet effective approach to improve the robustness of NMT to ASR noise by crafting ASR-specific noise training examples and incorporating the Chinese Pinyin feature as an additional input feature. Experiment results show that our method significantly outperforms the baseline and performs stably on three test sets with different amount of noise characters, while achieves a generalization improvement on a clean test set.

In future work, we would like to investigate appropriate methods to construct noise training examples for other types of ASR errors. Moreover, it is necessary to evaluate our approach on a realistic speech translation system.

\bibliography{ref}
\bibliographystyle{aaai}
\end{CJK}
\end{document}